\documentclass[sigconf,screen]{acmart}

\usepackage{booktabs}
\usepackage{multirow}
\usepackage{xspace}
\usepackage{microtype}
\usepackage{graphicx}
\usepackage{subcaption}
\usepackage{balance}
\usepackage{amsmath}
\usepackage{enumitem}
\usepackage{siunitx}
\usepackage{makecell}
\usepackage{listings}
\usepackage{pifont}
\usepackage{listings}
\usepackage{graphicx,xcolor,tabularx}
\usepackage[most]{tcolorbox}
\usepackage{ragged2e}

\definecolor{backcolour}{rgb}{0.95,0.95,0.92}

\definecolor{backcolour}{rgb}{0.95,0.95,0.92}
\definecolor{codegreen}{rgb}{0,0.6,0}
\definecolor{codegray}{rgb}{0.5,0.5,0.5}
\definecolor{codepurple}{rgb}{0.58,0,0.82}
\definecolor{codenavy}{rgb}{0.1,0.1,0.44}
\definecolor{codered}{rgb}{0.6,0,0}

\newcommand{\numvideos}{100}
\newcommand{\numhours}{41}
\newcommand{\numframes}{19{,}738}
\newcommand{\sumframes}{4{,}336{,}132}

\author{Wentao Lu}
\email{wlu4@ualberta.ca}
\affiliation{%
  \institution{University of Alberta}
  \city{Edmonton}
  \state{Alberta}
  \country{Canada}
}

\author{Alexander Senchenko}
\email{asenchenko@ea.com}
\affiliation{%
  \institution{Electronic Arts}
  \city{Vancouver}
  \state{British Columbia}
  \country{Canada}
}

\author{Alan Sayle}
\email{asayle@ea.com}
\affiliation{%
  \institution{Electronic Arts}
  \city{Guildford}
  \state{Surrey}
  \country{United Kingdom}
}

\author{Abram Hindle}
\email{abram.hindle@ualberta.ca}
\affiliation{%
  \institution{University of Alberta}
  \city{Edmonton}
  \state{Alberta}
  \country{Canada}
}

\author{Cor-Paul Bezemer}
\email{bezemer@ualberta.ca}
\affiliation{%
  \institution{University of Alberta}
  \city{Edmonton}
  \state{Alberta}
  \country{Canada}
}

\copyrightyear{2026}
\acmYear{2026}
\setcopyright{acmlicensed}
\acmConference[FSE '26]{Companion Proceedings of the 34th ACM Symposium on the Foundations of Software Engineering (FSE '26)}{June 5--9, 2026}{Montreal, Canada}
\acmBooktitle{Companion Proceedings of the 34th ACM Symposium on the Foundations of Software Engineering (FSE '26), June 5--9, 2026, Montreal, Canada}
\acmDOI{}
\acmISBN{}

\title{How Far Can VLMs Go for Visual Bug Detection?\\Studying \numframes{} Keyframes from \numhours{} Hours of Gameplay Videos}

\begin{document}

\begin{abstract}
Video-based quality assurance (QA) for long-form gameplay video is labor-intensive and error-prone, yet valuable for assessing game stability and visual correctness over extended play sessions. Vision language models (VLMs) promise general-purpose visual reasoning capabilities and thus appear attractive for detecting visual bugs directly from video frames. Recent benchmarks suggest that VLMs can achieve promising results in detecting visual glitches on curated datasets. Building on these findings, we conduct a real-world study using industrial QA gameplay videos to evaluate how well VLMs perform in practical scenarios. Our study samples keyframes from long gameplay videos and asks a VLM whether each keyframe contains a bug. Starting from a single-prompt baseline, the model achieves a precision of 0.50 and an accuracy of 0.72. We then examine two common enhancement strategies used to improve VLM performance without fine-tuning: (1) a secondary judge model that re-evaluates VLM outputs, and (2) metadata-augmented prompting through the retrieval of prior bug reports. Across \textbf{100 videos} totaling \textbf{41 hours} and \textbf{19,738 keyframes}, these strategies provide only marginal improvements over the simple baseline, while introducing additional computational cost and output variance. Our findings indicate that off-the-shelf VLMs are already capable of detecting a certain range of visual bugs in QA gameplay videos, but further progress likely requires hybrid approaches that better separate textual and visual anomaly detection.

\end{abstract}


\maketitle

\begin{figure}[ht!]
  \centering
  \begin{tcolorbox}[
      enhanced,
      width=\linewidth,
      colback=white,
      colframe=black,
      boxrule=0.8pt,
      arc=2pt, outer arc=2pt,
      drop fuzzy shadow,
      title=\bfseries Bug frame detected from a gameplay video,
      colbacktitle=black, coltitle=white,
      fonttitle=\bfseries\large,
      attach boxed title to top center={yshift=-2mm},
      boxed title style={boxrule=0.8pt, arc=2pt, outer arc=2pt},
      left=6pt, right=6pt, top=6pt, bottom=6pt, boxsep=5pt
    ]

    \centering
    \includegraphics[width=\linewidth]{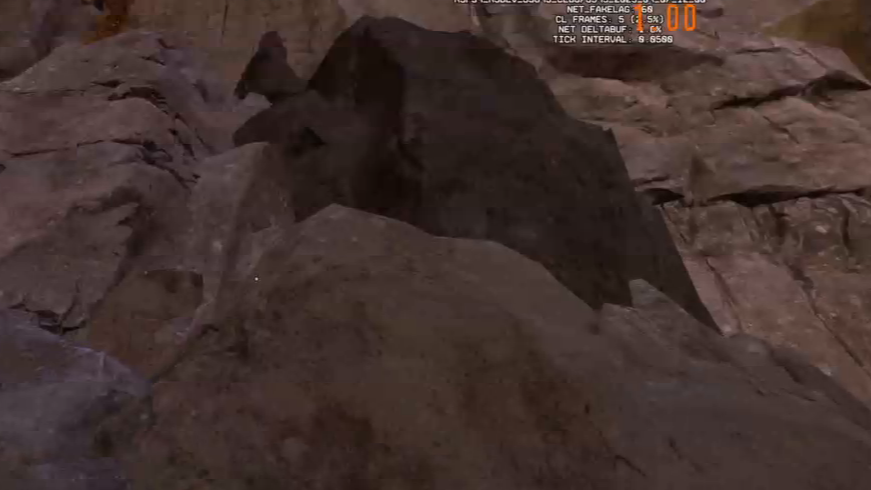}

    \tcblower
    \small
    \RaggedRight
    \setlength{\parskip}{4pt}

    \noindent\textbf{\textcolor{blue}{Bug label :}} True

    \noindent\textbf{\textcolor{codepurple}{Image description:}}
    The frame shows a first-person view in the game, aiming upward towards a rocky cliff face \ldots Importantly, the shadow cast on the large rock face in the center of the image seems irregular as it creates an unnatural dark rectangular patch that doesn't conform to the natural shape and topology of the rocks. This suggests the shadow mapping or rendering might be glitching, as real shadows would follow the contours of the rock more snugly rather than a flying rectangular shape \ldots\\[0.5ex]

    \noindent\textbf{\textcolor{red}{Bug description:}}
    The frame shows a clearly unnatural, hard-edged dark rectangular patch on the cliff face that does not follow the underlying rock geometry or lighting gradients; natural terrain shadows would wrap and soften around contours, whereas this shape looks like a misprojected shadow or failed shadow map \ldots\\[0.5ex]
  \end{tcolorbox}

  \caption{Example of a detected visual bug frame in a gameplay video using gpt-4.1-mini.}
  \label{fig:pipeline-example}
\end{figure}

\section{Introduction}
Large vision–language models (VLMs) have rapidly improved in open-ended visual understanding tasks~\cite{GPT4O, lu2025automatedbugframeretrieval, taesiri2025videogameqabenchevaluatingvisionlanguagemodels}. Their flexibility and low setup costs open up new opportunities for software quality assurance (QA), such as automatically analyzing long gameplay videos to identify visual bugs during automated QA tests.

A typical use case involves hundreds of hours of automated runs executed overnight to monitor for runtime exceptions or client instability (soak testing)~\cite{amey2022softwaretestdesign}. In practice, however, most of the footage is never reviewed end-to-end unless a critical error is suspected. As a result, these videos form a large but underused source of evidence about overall game stability and visual quality. In this report, we present our experience with re-purposing this existing video corpus for broader QA tasks, specifically, for detecting visible in-game anomalies using VLMs. We explore whether VLM analysis can capture visual bugs from these recordings without any additional model training. Although training data can be available, our goal is to evaluate the out-of-the-box effectiveness of general-purpose VLMs as plug-and-play components in an industrial QA workflow. Meanwhile, avoiding fine-tuning also reflects practical deployment needs where game builds change frequently, and maintaining up-to-date VLM knowledge through continuous retraining would be costly and operationally unsustainable in the long run, especially when newer model generations will inevitably replace older ones.

We therefore ask a simple question: \textbf{How effectively can VLMs analyze large volumes of gameplay video to detect visual bugs, and to what extent do Retrieval-Augmented Generation (RAG) or judge-based strategies improve their performance?} We study this empirically through a case study of \numvideos{} gameplay videos (\numhours{} hours with a total of of \sumframes{} frames).

Our study demonstrates that modern VLMs can turn previously underused gameplay recordings into a practical source of QA evidence and show that a simple prompting setup enables automated triage of long-form QA videos. This approach reduces the amount of video requiring manual inspection from millions of raw frames to only a few thousand candidate frames selected by VLMs.

Our main contributions are:
\begin{itemize}
  \item An experience-based case study demonstrating how gameplay recordings originally collected for stability testing can be re-purposed for large-scale visual QA through automated VLM analysis.
  \item A practical assessment of two common enhancement strategies, RAG and judge-based, applied to industrial video game QA data.
\end{itemize}

\begin{figure*}
    \centering
    \includegraphics[width=0.85\linewidth]{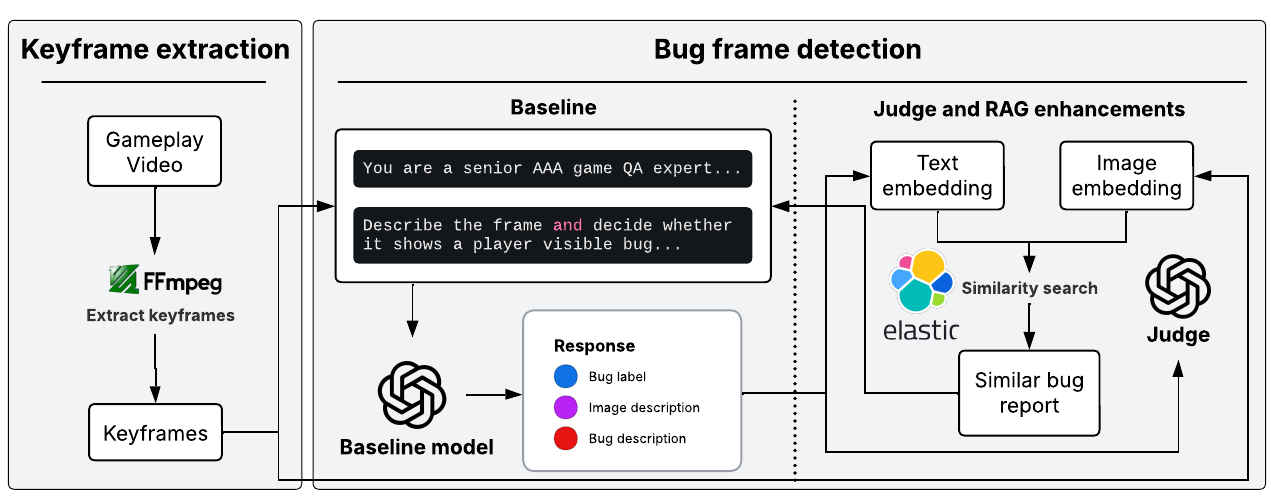}
    \caption{Overview diagram of our methodology}
    \label{fig:overview}
\end{figure*}

\section{Related Work}

\textbf{Traditional bug detection.}
Early studies in game quality assurance concentrated on bug detection through hand-crafted rules, visual and text features, or specialized machine learning models. Lin et al.~\cite{Dayi_game_bug_video} conducted one of the earliest studies, examining whether gameplay videos containing bugs could be automatically identified using publicly available metadata from online platforms. Building on this foundation,
Guglielmi et al. introduced \textit{GELID}~\cite{guglielmi2022usinggameplayvideosdetecting,guglielmi2023usinggameplayvideosdetecting}, a method that extracts bug moments from gameplay videos by leveraging video caption texts. Similarly, Truelove et al. proposed classifying video segments as buggy or non-buggy based on video captions~\cite{truelove2023findingneedlehaystackdetecting}. In subsequent research, Guglielmi et al. focused on detecting specific stuttering bug events and developed \textit{HASTE}~\cite{Stuttering}, which utilizes image similarity to identify stuttering occurrences.

In addition to caption-based methods, deep learning techniques have been employed to detect visual glitches directly from rendered frames. Ling et al. proposed a neural network approach to identify four types of rendered glitches in video games~\cite{ling2024usingdeepconvolutionalneural}, by training convolutional neural networks on labeled image data.
Recent work by Li et al.~\cite{yi2025hybridcofinetuningapproachvisual} explores visual bug detection using supervised machine learning with a hybrid co-finetuning approach that incorporates both labeled and unlabeled data.
These approaches generally depend on domain-specific training data and predefined bug categories, which restricts their ability to generalize across diverse games or to detect previously unseen visual anomalies.

\textbf{VLM-based bug detection.}
Recent advancements in large-scale vision–language models (VLMs) have opened new directions for visual bug detection. Compared to traditional approaches, VLM-based methods reduce the need for explicit ML training and manual annotation, making them suitable for a variety of visual bug detection tasks. 
Models such as GPT-4o demonstrate strong zero-shot understanding of complex scenes and can identify visual inconsistencies directly from raw images or video frames~\cite{GPT4O}.
Early work by Taesiri et al.~\cite{taesiri2022clipmeetsgamephysicsbug} demonstrated the effectiveness of using the Contrastive Language–Image Pre-Training (CLIP) model to identify game videos containing bugs through text queries, and language models are good as a zero-shot bug detector. They later
introduced \textit{VideoGameBunny}~\cite{Taesiri_VideoGameBunny}, a model fine-tuned on extensive video game image datasets and associated instruction pairs,
demonstrating an improved understanding of the game context and being more accurate in-game responses.
Their subsequent studies introduced GlitchBench~\cite{taesiri2024glitchbenchlargemultimodalmodels} and VideoGameQA-Bench~\cite{taesiri2025videogameqabenchevaluatingvisionlanguagemodels}, which established benchmarks for evaluating VLMs on a range of game quality assurance tasks. These benchmarks demonstrated strong performance in identifying visual glitches, particularly in needle-in-a-haystack tasks that require locating bug instances within gameplay videos.
Building on this line of research, our previous work~\cite{lu2025automatedbugframeretrieval} utilized VLMs to detect potential bug frames in gameplay videos by integrating bug descriptions from reports with keyframes extracted from videos.

\textbf{Unlike prior studies} that primarily rely on public or synthetically generated datasets, \textbf{our work builds upon industrial, real-world QA gameplay footage}, a continuation from our previous study~\cite{lu2025automatedbugframeretrieval}. We adapt the used keyframe-based strategy for extended long gameplay video by evaluating VLM performance on production-scale, long-duration gameplay recordings that are automatically collected over hundreds of hours, and by exploring methods to enhance detection accuracy without fine-tuning.  
Specifically, two complementary techniques are investigated: RAG and VLM as a judge model.

\begin{table*}
\caption{Precision and accuracy across different variants and judge models}
\centering
\begin{tabular}{l c c c c c c}
\toprule
 & \multicolumn{3}{c}{\textbf{Precision}} & \multicolumn{3}{c}{\textbf{Accuracy}} \\
\cmidrule(lr){2-4}\cmidrule(lr){5-7}
4.1-mini (Baseline)        & \multicolumn{3}{c}{0.50} & \multicolumn{3}{c}{0.72} \\
5-mini        & \multicolumn{3}{c}{0.35} & \multicolumn{3}{c}{0.57} \\
\midrule
\textbf{Enhancements} & \textbf{4.1-mini} & \textbf{4o-mini} & \textbf{5-mini} & \textbf{4.1-mini} & \textbf{4o-mini} & \textbf{5-mini} \\
\cmidrule(lr){2-4}\cmidrule(lr){5-7}
+ Judge                        & 0.47 & 0.48 & 0.55 & 0.72 & 0.71 & 0.74 \\
+ Image RAG           & 0.37 & 0.35 & 0.46 & 0.67 & 0.67 & 0.71 \\
+ Text RAG (Bug description)    & 0.47 & 0.47 & 0.57 & 0.71 & 0.71 & 0.75 \\
+ Text RAG (Image description) & 0.45 & 0.45 & 0.54 & 0.70 & 0.70 & 0.74 \\
\bottomrule
\end{tabular}
\label{tab:overall}
\end{table*}

\begin{table}
\centering
\caption{Distribution of primary bug categories (randomly sampled 1,000 results from baseline performance)}
\label{tab:glitch-category-dist}
\begin{tabular}{l S[table-format=2.2]}
\toprule
\textbf{Category}  & \textbf{Percentage (\%)} \\
\midrule
\textbf{Log message(s)}  & 46.80 \\
\midrule
\textbf{Other visual anomalies} & 53.20 \\
- Rendering \& Texture   & 18.60 \\
- UI \& HUD              & 18.10 \\
- Animation \& VFX       &  6.10 \\
- Lighting \& Shadow     &  5.90 \\
- Physics \& Collision   &  3.10 \\
- Game Crash \& Logic    &  0.70 \\
- Performance            &  0.60 \\
- Other                  &  0.10 \\
\bottomrule
\end{tabular}
\end{table}

\vspace{-12mm}
\section{Methodology}
\label{sec:methodology}
This section outlines our approach for detecting visual bugs using VLMs and keyframes extracted from long gameplay videos. 
Figure~\ref{fig:overview} illustrates the steps of our approach. For each long gameplay recording, we (1) extract keyframes, (2) present each to a VLM, and (3) collect the model’s decision on whether a bug is present.

\subsection{Keyframe extraction}
\textit{\textbf{Keyframes capture major visual changes in the video, enabling analysis to focus on the most representative moments of gameplay}.} Long QA test recordings often span tens of minutes or hours, resulting in substantial redundancy across adjacent frames. Analyzing every frame is computationally intensive and unnecessary for detecting visible anomalies. Building on our previous work~\cite{lu2025automatedbugframeretrieval}, keyframes are extracted using FFmpeg~\cite{FFmpeg} to significantly reduce the number of frames processed while preserving those most likely to contain visual anomalies. This sampling strategy achieved 98.79\% bug coverage in prior evaluations, and in this study, we evaluated the same First Person Shooter (FPS) game.

\vspace{-3mm}
\subsection{Bug frame detection}
The bug frame detection task is formulated as a single binary decision accompanied by a brief justification: given a keyframe, the model must classify it as buggy or normal and provide a rationalization in one or two sentences. This task instruction remains consistent across all experiments. Subsequent techniques only (1) re-evaluate the model's answer or (2) modify the information provided to the model. To enhance VLM performance in this task, two complementary strategies are incorporated, motivated by recent advances in model evaluation and context enrichment:

\textbf{Judge models:} We introduce secondary VLMs to re-evaluate the primary output.
Previous 'LLM-as-a-Judge' studies have demonstrated that models can serve as reliable evaluators, exhibiting high agreement with human judgments~\cite{verga2024replacingjudgesjuriesevaluating, li2024llmsasjudgescomprehensivesurveyllmbased, LLMasJudgeGuildline, chen2024mllmasajudge}.
In this study, the judge model receives the frame, the decision label, and its rationale, and then either confirms or overturns the decision.
This verification is primarily designed to reduce false positives in bug detection.

\textbf{Metadata-augmented prompting using RAG: }To expand the model’s knowledge base without fine-tuning, prior bug reports most similar in visual or textual content are retrieved and appended to the prompt context~\cite{liu2024raghelpreasoningllm, ding2025slowfastvadvideoanomalydetection, bdcc8090115, bonomo2025visualragexpandingmllm}.
Providing metadata and textual descriptions of expected behavior has been shown to improve VLM reasoning regarding anomalies~\cite{macklon2025exploringcapabilitiesvisionlanguagemodels,lu2025automatedbugframeretrieval}.
In practice, for each keyframe, a similar bug report is retrieved from the database and provided alongside the task instruction. The model then issues the same binary decision with a brief justification.

\subsubsection{Baseline}
The baseline serves as the simplest reference pipeline. Each extracted keyframe is directly presented to the VLM (gpt-4.1-mini~\cite{4.1mini}) using the above task instruction, producing:
\begin{itemize}
    \item \textbf{Bug label (True/False):} Whether the frame contains a visible bug.
    \item \textbf{Image description:} A detailed description of the keyframe.
    \item \textbf{Glitch description:} A short rationale explaining the observed bug, if labeled True.
\end{itemize}

This baseline reflects the model’s out-of-the-box ability to generalize to the game QA domain. 

\subsubsection{Judge and RAG enhancements}
With the goal of improving the baseline performance, we first apply the judge step directly over the baseline results. Subsequently, we augmented the baseline approach with an additional retrieval-augmented prompting step. Three complementary retrieval strategies are used to identify the most relevant prior bug reports for each keyframe:

\begin{enumerate}
    \item \textbf{Image similarity:} The keyframe is embedded using a CLIP image encoder~\cite{clip-vit-base-patch32} and compared (based on cosine similarity between two vectors) against stored image embeddings of historical images attached to bug reports. 
    \item \textbf{Image description similarity:} The description of the keyframe produced by the baseline model is embedded using a text embedder and matched against historical bug report descriptions to capture semantic similarity. Unlike the next strategy, this gives all keyframes a chance to retrieve similar reports regardless of any potential bugs.
    \item \textbf{Bug description similarity:} When the baseline model identifies a bug (label is True), its generated bug description is embedded and compared to historical report texts to retrieve examples with similar bugs.
\end{enumerate}


Finally, we re-run the task using an updated prompt that includes (1) the keyframe image, (2) the baseline model’s initial decision and rationale, and (3) the retrieved bug report. 

\vspace{-2mm}
\section{Experimental setup}
\label{sec:experiment_setup}
\textbf{Gameplay video dataset:}
Our evaluation is conducted on a corpus of 100 long gameplay test videos drawn at random from daily automated runs within an industrial QA pipeline. These videos comprise a total of 41 hours of footage, encompassing 4,336,132 video frames, with a mean video length of 24 minutes and 5 seconds. Segmentation using FFmpeg~\cite{FFmpeg}, consistent with our prior work~\cite{lu2025automatedbugframeretrieval}, yields 19,738 keyframes. To establish a ground truth, all extracted keyframes were manually reviewed and labelled by one of the authors. Each frame is annotated with a binary label indicating the presence or absence of a player-visible bug.

\textbf{Bug categories:} Unlike regular gameplay recordings, automated QA test runs constantly display on-screen log messages generated by backend logging systems. When these messages indicate critical errors or failures, they are considered bugs. Consequently, we introduced an additional category to our study beyond the seven categories used in our previous work~\cite{lu2025automatedbugframeretrieval}, shown in Table \ref{tab:glitch-category-dist}.

\textbf{Bug reports:}
In addition to this labeled gameplay video corpus, we utilize a database of historical bug reports collected from the internal JIRA system, a widely adopted platform for bug reports~\cite{Jira}. Each report includes a bug description, optional screenshot artifacts, and associated developer metadata. The database is pre-vectorized for internal services: text is embedded using \textit{gte-qwen2-1.5b-instruct-embed-f16}~\cite{Qwen2-1.5B}, and images are embedded using \textit{clip-vit-base-patch32}~\cite{clip-vit-base-patch32}. Embeddings are stored in an ElasticSearch (ES) index~\cite{elastic}. This vectorized database enables similarity search queries based on cosine similarity and underpins the RAG prompting strategies evaluated in Section~\ref{sec:results}. Due to legal constraints, we cannot publish our data or provide a more detailed description of the embedding strategy.
  
\textbf{Used models:}
All judge and baseline models used in this study are sourced from OpenAI~\cite{OpenAI},  accessed through API version ``2025-04-01-preview" (the latest version at the time)~\cite{Azure}, which provides industry-leading, production-grade VLMs. This ensures that model behavior and evaluation results are based on standardized, widely recognized architectures. We specifically select the \textit{mini} variants to balance performance and cost, reflecting practical considerations for deployment in large-scale, production-ready QA environments. Due to internal access restrictions, we are limited to using OpenAI models.

\textbf{Performance metrics:}  All metrics are reported as per-video means to ensure that longer recordings with more keyframes do not disproportionately influence the overall results. We focus on two key metrics: \textbf{precision} (fraction of detected bugs that are actual bugs) and \textbf{accuracy} (fraction of correct detection overall).

\section{Results}
\label{sec:results}


\subsection{Baseline performance}
\textit{Motivation: }Prior to evaluating enhancement strategies, the performance of a VLM on the bug detection task using only visual input is evaluated. This baseline establishes the minimum performance achievable without external guidance.

\textit{Approach: }Each extracted keyframe is analyzed by the VLM using the prompt design detailed in Section~\ref{sec:methodology}. The gpt-4.1-mini model is selected as the baseline due to its high ranking (\#3) in the previous \textit{VideoGameQA Bench} study~\cite{taesiri2025videogameqabenchevaluatingvisionlanguagemodels} and its cost efficiency.
For comparison, we also evaluate the same baseline configuration using 5-mini, but omit 4o-mini due to its significantly higher cost (approximately 20x more tokens than 4.1-mini, as a known issue to OpenAI).

As shown in Table~\ref{tab:overall}, the baseline achieves a precision of 0.50 and an accuracy of 0.72 across \numvideos{} gameplay videos.  
These results indicate that \textbf{off-the-shelf VLMs generalize effectively to game QA videos}. With a precision of 0.50, half of the model-flagged frames correspond to true bugs. Reviewing only the flagged frames reduces human inspection from \sumframes{} total frames to 2,002 ($\approx$0.05\%), thereby converting recordings into a manageable triage queue with candidate bug descriptions generated by the model.

In addition to performance metrics, the types of visual bugs that the VLM is capable of detecting are examined.
Table~\ref{tab:glitch-category-dist} presents the distribution of primary detected bug categories from baseline results.
Nearly half of the identified bugs (\textbf{46.8\%}) are on-screen error messages generated by backend logging systems. These overlays typically indicate lower-level system or engine errors, such as missing assets, null references, or network failures, rather than visual anomalies.
However, such text overlay bugs are readily detected by the model, indicating that VLMs can effectively recognize interface overlays to diagnostic and prioritized error messages.
However, these cases could potentially be addressed more efficiently using traditional optical character recognition (OCR), which would allow VLMs to focus on non-textual anomalies.
The remaining \textbf{53.2\%} of detected bugs comprises a diverse range of visual anomalies.
The VLM can identify many of these cases based solely on visual context. Notably, visual bugs within HUD \& UI often persist across multiple consecutive keyframes, resulting in duplicate detections.
A promising direction for future research is to cluster bug keyframes based on visual similarity or bug type,  merging redundant detections and thus effectively capturing recurring or persistent visual anomalies.

\vspace{5pt}
\noindent
\fbox{\parbox{\columnwidth}{
  \textit{An off-the-shelf VLM can achieves 50\% precision while reducing manual review to a manageable subset of frames, and discover bugs that would otherwise remain unseen in unreviewed QA recordings.}
}}

\subsection{Judge and RAG performance}

\subsubsection{Judge models}  
In this setting, the judge model receives the keyframe, the baseline predicted label, and the associated reasoning, and then re-evaluates the decision.

Overall, \textbf{judge models yield mixed outcomes.}  
Both 4.1-mini and 4o-mini perform slightly below the baseline (precision 0.47–0.48, accuracy 0.71–0.72), whereas 5-mini demonstrates a modest improvement, achieving 0.55 precision and 0.74 accuracy.
However, when 5-mini is used directly as the baseline detector model instead of 4.1-mini, its performance is substantially lower (0.35 precision, 0.57 accuracy). This result suggests that 5-mini is more effective as an evaluator than as a predictor.
\vspace{-2mm}

\subsubsection{Retrieval-augmented prompting}  

Results indicate that \textbf{RAG often introduces noise}.
Image-based retrieval decreases performance (precision = 0.38, accuracy = 0.68) because visually similar images are often retrieved, yet the corresponding bug report descriptions are unrelated.
Text-based retrieval yields improved performance, particularly when retrieved using bug description similarity and reviewed by 5-mini, achieving 0.57 precision and 0.75 accuracy. This result slightly surpasses both the baseline and the judge-only technique.

Both judge and RAG steps double inference time and cost compared to the baseline, since they require a second model call and additional embedding retrieval queries.
Given the marginal accuracy gains, this highlights a trade-off between performance improvement and practical deployability in QA pipelines.

\vspace{5pt}
\noindent
\fbox{\parbox{\columnwidth}{
  \textit{Judge and RAG strategies yield only marginal, model-specific gains. A capable judge can refine baseline outputs, whereas text-based retrieval can sometimes improve results.}
}}

\section{Discussion}
\textbf{Comparison to prior work:}
Taesiri et al. introduced \textit{VideoGameQA-Bench}~\cite{taesiri2025videogameqabenchevaluatingvisionlanguagemodels}, which features the challenging Needle-in-a-Haystack (NIAH) task.
In this task, a model is required to identify anomalous frames and their corresponding timestamps in a video, referred to as “needles,” which are among thousands of normal frames sampled uniformly at one frame per second.
Within this framework, 4.1-mini achieved an accuracy of only 10\%.
Our study constitutes a variant of the NIAH task. Instead of uniform sampling, our analysis focuses on extracted keyframes that capture major visual scenes and substantially reduce redundancy. Second, our industrial QA dataset differs qualitatively. A large portion of detected anomalies are on-screen log messages, which are visually easily recognized by the model. Finally, our formulation analyzes each keyframe, similar to image-based detection in \textit{VideoGameQA-Bench} that scores 76.8\% in accuracy.
Bug frames remain rare within these keyframe sequences, and our result achieves 74\% accuracy, substantially outperforming the NIAH baseline but comparable to image-based glitch detection.
Additionally, the use of keyframes preserves timestamp metadata from the original videos, enabling direct bug localization without requiring the model to infer.

\vspace{-2mm}
\section{Threats to Validity}
\label{Sec:threats}

\textbf{Internal validity:} A potential threat is the ground truth labeling process. All keyframes were labeled by a single author, which risks annotation mistakes and subjective bias. Some label noise is likely present and may affect the performance evaluation.

\textbf{External validity:} Our dataset consists of long QA recordings from a single game within one industrial pipeline.
Other game genres or earlier development stages may exhibit different bug types and visual styles, which could change VLM behaviour.

Another threat is that our analysis is limited to OpenAI mini VLM variants because of internal access constraints.
Other model families with larger-capacity models might exhibit different performance and different sensitivity to RAG and judge strategies.
Similarly, our RAG setups depend on one specific embedding configuration and JIRA-based report structure. Different bug tracking systems or embedding strategies might yield different RAG quality.

\vspace{-2mm}
\section{Conclusion}
\label{sec:conclusion}

In this study, we evaluated the capability of large vision-language models to detect visual bugs directly from industrial QA gameplay videos. Using a dataset of {\numvideos} videos, {\numhours} hours and {\numframes} keyframes, we showed that a simple baseline with gpt-4.1-mini 
can reduce manual inspection of entire videos to a manageable triage queue of frames.

We further examined two widely used strategies for improving large model performance: judge-based re-evaluation and RAG.
Across the configurations we evaluated, image-based retrieval degraded performance, while text-based retrieval yielded only marginal gains.
Similarly, a stronger judge (gpt-5-mini) could refine baseline results, but improvements were modest.

Overall, these findings highlight the practical potential of VLM-based triage for industrial QA workflows, substantially reducing the volume of video that requires human review. However, our findings also show that improving VLM performance is not trivial. 
Further progress will require more targeted hybrid designs, such as separating textual OCR, incorporating temporal reasoning over sequences of frames, and clustering redundant bug keyframes to better exploit the structure of QA gameplay recordings.

\bibliographystyle{ACM-Reference-Format}
\bibliography{references}
\end{document}